\documentclass[sigconf]{acmart} 
\AtBeginDocument{%
  }

\setcopyright{acmlicensed}
\copyrightyear{2025}
\acmYear{2025}
\setcopyright{acmlicensed}\acmConference[IVA Adjunct '25]{ACM International Conference on Intelligent Virtual Agents}{September 16--19, 2025}{Berlin, Germany}
\acmBooktitle{ACM International Conference on Intelligent Virtual Agents (IVA Adjunct '25), September 16--19, 2025, Berlin, Germany}
\acmDOI{10.1145/3742886.3756708}
\acmISBN{979-8-4007-1996-7/2025/09}

\begin{document}

\title{Spotter+GPT: Turning Sign Spottings into Sentences with LLMs}

\author{Ozge Mercanoglu Sincan}
\affiliation{%
\institution{CVSSP}
  \institution{University of Surrey}
  \city{Guildford, Surrey}
  \country{United Kingdom}
}
\email{o.mercanoglusincan@surrey.ac.uk}

\author{Richard Bowden}
\affiliation{%
  \institution{CVSSP}
  \institution{University of Surrey}
  \city{Guildford, Surrey}
  \country{United Kingdom}
}
\email{r.bowden@surrey.ac.uk}

\renewcommand{\shortauthors}{Sincan et al.}

\begin{abstract}
Sign Language Translation (SLT) is a challenging task that aims to generate spoken language sentences from sign language videos. In this paper, we introduce a lightweight, modular SLT framework, Spotter+GPT, that leverages the power of Large Language Models (LLMs) and avoids heavy end-to-end training. Spotter+GPT breaks down the SLT task into two distinct stages. First, a sign spotter identifies individual signs within the input video. The spotted signs are then passed to an LLM, which transforms them into meaningful spoken language sentences. Spotter+GPT eliminates the requirement for SLT-specific training. This significantly reduces computational costs and time requirements. The source code and pretrained weights of the Spotter are available online \footnote{\url{https://gitlab.surrey.ac.uk/cogvispublic/sign-spotter}}.

\end{abstract}


\begin{teaserfigure}
  \includegraphics[width=\textwidth]{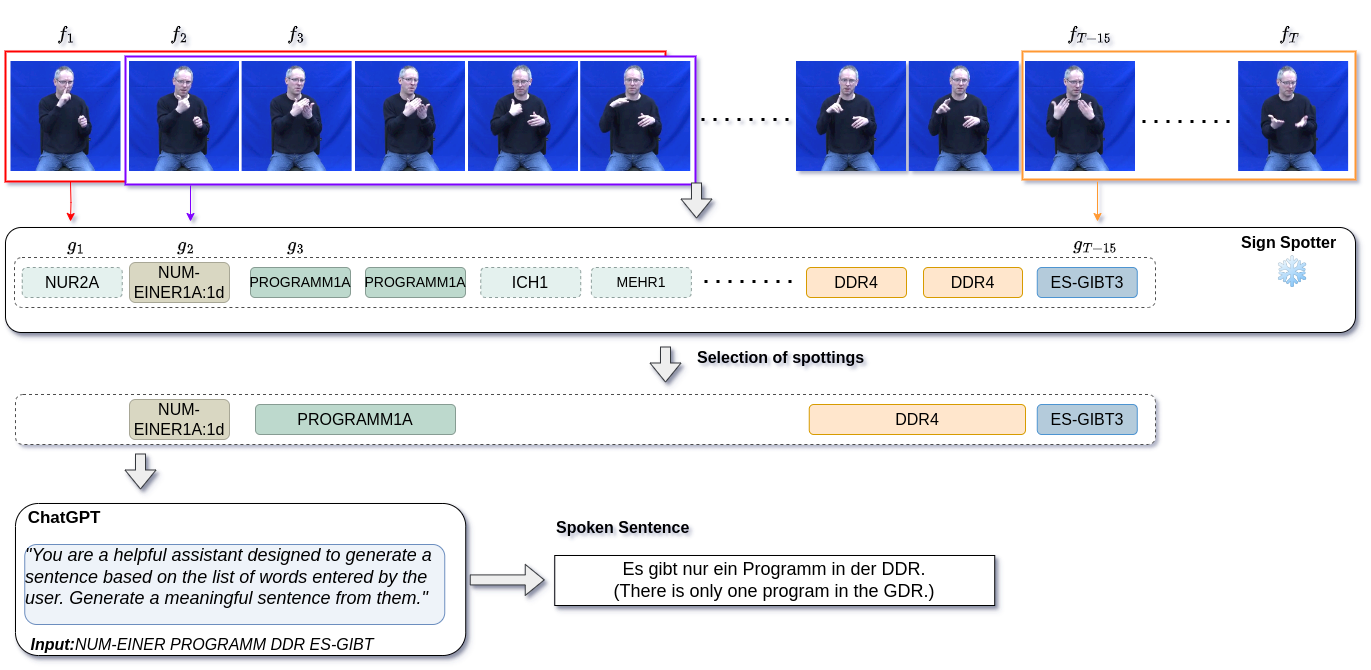}
  \caption{Overview of the proposed sign language translation framework. The system spots signs by processing video input and generates spoken language sentences via ChatGPT.}
  \label{fig:method}
\end{teaserfigure}


\maketitle
   
\section{Introduction}
\label{sec:intro}
Sign languages are visual languages that rely on manual hand articulations, facial expressions, and body movements. To bridge communication gaps between the Deaf community and hearing people, Sign Language Translation (SLT) (sign language $\rightarrow$ spoken language) and Sign Language Production (SLP) (spoken language $\rightarrow$ sign language) systems hold great significance. 

\begin{figure}[h!]
  \centering
  \includegraphics[width=\linewidth]{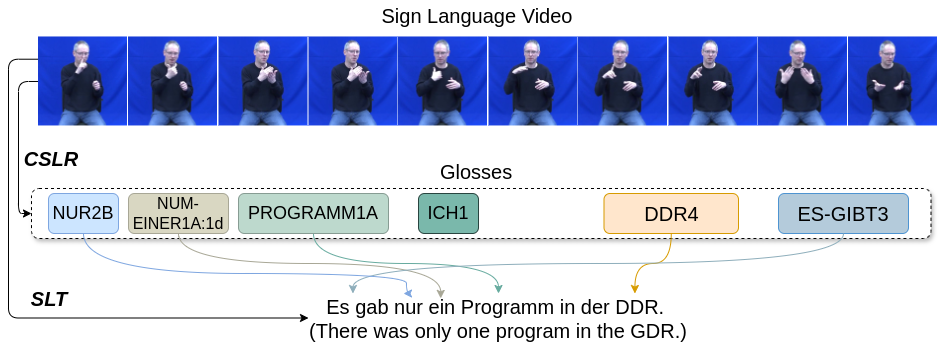}
  \caption{Overview of the CSLR (Continuous Sign Language Recognition) and SLT (Sign Language Translation) tasks.}
  \label{fig:slt}
\end{figure}

SLT is often formulated as a Neural Machine Translation (NMT) task since it aims to generate spoken/written language sentences from sign language videos \cite{Camgoz_2018_CVPR}. Compared to classical text-based NMT approaches, which work on easily tokenizable text, SLT deals with continuous sign language videos, which are hard to align and tokenize. Furthermore, spoken and sign languages have different grammar, and the order of the sign glosses and spoken language words are different as seen in Fig. \ref{fig:slt}. 

To address this issue, some researchers have approached SLT as a combination of two sub-tasks; Sign Language Recognition (SLR), recognizing the constituent signs of sequences and then translating the recognized signs into meaningful spoken language sentences \cite{Camgoz_2018_CVPR, zhang2022sltunet, camgoz2020sign}. Some researchers approach the first task as Continuous Sign Language Recognition (CSLR) and try to detect sequences of glosses as an intermediate representation to represent sign language videos. Camgoz et. al \cite{Camgoz_2018_CVPR, camgoz2020sign} showed that using gloss-based intermediate representations improved SLT performance significantly. The common approach in CSLR is learning spatial and temporal visual representations with a sequence-to-sequence Connectionist Temporal Classification (CTC) loss \cite{graves2006connectionist}. These CSLR approaches require gloss supervision, which is hard and time-consuming to accurately annotate.

On the other hand, some researchers tackle SLT in an end-to-end manner by training SLT models that produce spoken language sentences directly from sign language videos \cite{camgoz2020sign, yao2023sign, Zhou_2021_CVPR}. Although end-to-end approaches can achieve excellent results on small SLT datasets, such as PHOENIX-2014-T \cite{Camgoz_2018_CVPR}, they under perform on SLT datasets that are weakly aligned or have a large domain of discourse \cite{sincan2023context, albanie2021bbc, muller2022findings}.

In this work, we explore an alternative, modular approach: Spotter+GPT. Our goal is to reduce the cost and complexity of SLT pipelines by leveraging pre-trained Large Language Models (LLMs) instead of training a gloss-to-text translation model. We first train a sign spotter using a large sign language dataset from the linguistic domain. Then, we utilize our spotter to recognize the sequence of sign glosses from continuous sign language videos. These glosses are then passed to the LLM via prompting to generate spoken sentences. We evaluate our method on the MeineDGS dataset and a custom DGS-20 benchmark and compare it to traditional gloss-to-text transformer baselines. This pipeline does not require training an SLT-specific decoder and can generate semantically coherent translations, particularly in controlled settings.

\section{Related Work}
\label{sec:relatedwork}

\subsection{Sign Language Recognition}
SLR can be divided into two categories; Isolated Sign Language Recognition (ISLR) \cite{joze2018ms, li2020word, albanie2020bsl, sincan2020autsl, albanie2021bbc, wong2022hierarchical, bohacek2023learning} and Continuous Sign Language Recognition (CSLR) \cite{pu2019iterative, zheng2023cvt, wei2023improving, cui2019deep}. ISLR focuses on recognizing a single sign from the video, while CSLR focuses on recognizing the sequence of glosses. 
CSLR is a weakly supervised recognition task since continuous SLR datasets \cite{Camgoz_2018_CVPR, duarte2021how2sign} usually provide gloss sequences without explicit temporal boundary since labeling each gloss frame by frame is a time-consuming process. For this reason, using the CTC loss \cite{graves2006connectionist} became popular \cite{camgoz2020sign, pu2019iterative, zheng2023cvt, wei2023improving}.

Some researchers develop CSLR models with the assistance of an ISLR model \cite{cui2019deep, wei2023improving}. Cui et. al \cite{cui2019deep} first trained a feature extraction module and then finetuned the whole system iteratively. Wei et. al \cite{wei2023improving} trained two ISLR models in two different sign languages and then proposed a multilingual CSLR by utilizing cross-lingual signs with their assistance. In this work, we utilized an ISLR model to detect the sequence of glosses.

\subsection{LLMs for Sign Language Translation}
The development of Large Language Models (LLMs) has led to significant improvements in natural language processing tasks \cite{radford2018improving, brown2020language, touvron2023llama}. After OpenAI introduced ChatGPT, which was trained using Reinforcement Learning from Human Feedback (RLHF), it has attracted significant attention because of its success in several tasks, including question answering, summarization, machine translation, etc \cite{peng2023towards, qin2023chatgpt, manakhimova2023linguistically}. 

Despite the effectiveness of ChatGPT, only a few studies have explored ChatGPT in the context of sign language. Shahin and Ismail \cite{shahin2023chatgpt} evaluated ChatGPT's capabilities on gloss-to-text and text-to-gloss translations with a limited set of phrases, such as with only 5 medical-related statements in English and Arabic sign-spoken language pairs. While promising, their experiments were minimal and relied on given gloss inputs rather than real video inputs. They found that it performs better when translating to English rather than Arabic.

More recently, several studies have begun integrating LLMs into SLT pipelines, often through training adapters or fine-tuning \cite{gong2024llms, wong2024sign2gpt, chen2024factorized, jang2025lost, liang2024llava}. However, such systems typically require additional training and resources, limiting their ease of deployment.

In contrast, our approach, Spotter+GPT, is a lightweight SLT framework that uses a pretrained sign spotter to detect glosses from video, and directly prompts ChatGPT for spoken sentence generation. To the best of our knowledge, it is the first to combine gloss spotting from raw video with prompt-based LLM translation, without requiring any additional training or adaptation for translation.

\section{Method}
\label{sec:method}

As shown in Fig. \ref{fig:method}, our approach contains a two-step process. First, we train a sign spotter to identify individual signs. This model was trained to recognize a single isolated sign from the input video. Once trained, the spotter is applied to continuous sign language videos in a sliding window manner to get gloss predictions. Then we pool and threshold these detections to obtain the final set of identified glosses. We call this first step Sign Spotting (Section \ref{section-spotter}). Subsequently, we employ a pretrained LLM to convert these sequences of glosses into spoken language sentences. To achieve this, we prompted a ChatGPT model (Section \ref{section-gpt}).

\subsection{Sign Spotting}
\label{section-spotter}
As spotter, we employ an I3D model \cite{carreira2017quo}. We prepare isolated sign sequences from a continuous sign language dataset, MeineDGS \cite{konrad2019meine}, where the glosses are annotated in frame level. Within the MeineDGS dataset the average duration of a gloss is 10 frames while we train our I3D models with a window size of 16. If an isolated sign sequence is shorter than 16 frames, we repeat the last frame. For longer sign instances we generate multiple samples by using a sliding window with a stride of 8.

In the training time, the model takes 16 consecutive frames. We resize the input images to 256 $\times$ 256 and then crop to a 224 $\times$ 224 region. We replaced the ReLU activation function with the Swish activation function \cite{ramachandran2017searching} as it improves sign language recognition performances \cite{jiang2021skeleton, wong2022hierarchical}.

Following \cite{albanie2021bbc}, we utilize cross-entropy loss and an SGD optimizer  \cite{sutskever2013importance} with momentum 0.9, batch size 4, and an initial learning rate of 0.01. We decreased the learning rate by a factor of 10 when validation loss plateaus for 4 consecutive epochs. We use label smoothing of 0.1 to prevent overfitting. While our input videos are cropped randomly during training time, it is cropped from the center during evaluation. We applied color augmentation during training.

After training our I3D model, we employ it to spot signs in coarticulated continuous sign language videos in a sliding window manner. With a stride of 1, for the given input video with $T$ frames, we obtain $T-15$ gloss predictions. Then, we propose a straightforward yet efficient solution to produce a final sequence of glosses. First, we filter gloss predictions by a threshold, based on the model's prediction confidence. Following this, we collapse the consecutive repeated predictions to obtain the final sequence of gloss predictions. 

\subsection{ChatGPT}
\label{section-gpt}

Due to the success of ChatGPT in many tasks, we utilize it to create spoken sentences from glosses. 
This strategy not only eliminates the requirement for gloss-to-text model training but also offers potential advantages. ChatGPT can generate fluent and context-aware sentences, making it a strong candidate for translating glosses into coherent sign language sentences. 

It is worth noting that the order of the glosses in sign languages are different from the word order in spoken languages. Meanwhile, the vocabulary of our gloss spotter is constrained by the number of classes on which the I3D model was trained on. However, the dynamic nature of ChatGPT allows a flexible translation process, enabling us to overcome the limitations of a fixed gloss set and, different ordering.

We prompt `gpt-3.5-turbo' version of ChatGPT  via the OpenAI Python API without fine-tuning it. For each input video, after we obtain a series of glosses, we pass them to ChatGPT and prompt the model to create a meaningful sentence using those glosses.

The spotter is capable of real-time inference on systems equipped with a GPU. We tested it on two machines: a desktop with an NVIDIA RTX 3090 and a laptop with an NVIDIA RTX 5000. In both cases, the system operated smoothly, achieving a minimum of 25 frames per second. Since ChatGPT is accessed via an external API, its response time depends on network conditions. In our observations, the typical latency per prompt was approximately 1-2 seconds.

\subsection{Prompt Engineering}
We initially set our prompt by only defining our task as generating a sentence based on the list of words entered by the user. However, we observed that sometimes ChatGPT produces outputs like ``Sorry, I could not generate a sentence." in cases when spotter failed to detect any gloss or when the number of detected glosses was insufficient. To address this, we refined the prompt by adding two explicit rules to avoid unrelated sentences and to produce ``No translation." in cases where a meaningful sentence could not be generated. The final prompt is as follows:

\begin{itemize}
    \item You are a helpful assistant designed to generate a sentence based on the list of words entered by the user. You need to strictly follow these rules: 
    \begin{enumerate}
               \item The user will only give the list of German words separated by a space, you just need to generate a meaningful sentence from them. 
        \item Only provide a response containing the generated sentence. If you cannot create a German sentence then respond with ``No Translation". 
    \end{enumerate} 
\end{itemize}

\section{Experiments}
\label{sec:experiments}

\subsection{Dataset and Preprocessing}

\textbf{MeineDGS.} It is a large linguistic German Sign Language (DGS) dataset \cite{konrad2019meine}. Videos contain free-flowing conversation between two deaf participants. We follow the sign language translation protocol set \cite{saunders2022signing} on MeineDGS, which has 40,230 training, 4,996 development, and 4,997 test sentences. 

We use the MeineDGS-V split \cite{saunders2022signing}, which distinguishes between sign variants, each containing the same meaning but with differing motions. MeineDGS-V has approximately 10,000 glosses available for training. However, some signs have relatively few examples or some of them are singletons. To make the dataset more balanced for isolated SLR model training, we selected glosses that have more than 12 samples and we exclude the \textit{INDEX} gloss as it stands for a pointing, making it the predominant gloss in the dataset. This criterion leads to 2,301 classes. We train our I3D spotter on these 2,301 classes using train, validation, and test splits of Saunders et. al \cite{saunders2022signing}. 

Please note that in \cite{saunders2022signing} the spoken sentences are lowercase, punctuation was removed, and all German characters (ä, ö, ü, and ß) were replaced with corresponding English letters. This will cause some words to change meaning. Therefore, instead of the sentence representation of \cite{saunders2022signing}, we used the original spoken language sentences \cite{konrad2019meine} in the sign language translation task.

\textbf{DGS-20 Videos.} To further evaluate our approach, we collected a small dataset containing 20 German Sign Language (DGS) videos. These videos were recorded by a single deaf signer, who performed 10 unique sentences, each repeated twice for consistency and variation. The glosses used in these sentences were selected from the 2,301-class vocabulary of our trained sign spotter, ensuring full coverage. Some example sentences are presented in Table~\ref{tab:dgs20}.

\begin{table}[!h]
 \caption{Some examples from DGS-20 dataset. }
	\centering
	\begin{tabular}{p{0.2cm}p{7.3cm}}
		\hline 
    \textbf{ID} &  \textbf{German Sentence (English Translation)}  \\ \hline 
        1 & Die Familie isst abends im Restaurant.  \\   
          & (The family eats in the restaurant in the evening.) \\

        2 & Der Dolmetscher spricht mit der Familie des Mädchens. \\ 
        & (The interpreter speaks to the girl's family.) \\

        3 & Der Dozent fährt morgens mit dem Fahrrad zur Universität.  \\   
        & (Tomorrow the lecturer will ride his bike to the university.)\\

        4 & Der Schauspieler spricht viel über Politik und Kultur. \\
         & (The actor talks a lot about politics and culture.) \\
         \hline
        \end{tabular}
    \label{tab:dgs20}
	\end{table}

\subsection{Evaluation Metrics}
 We use BLEU \cite{papineni2002bleu}, and BLEURT \cite{sellam2020bleurt} metrics to evaluate the performance of our SLT approach. BLEU is a metric based on the precision of n-grams (consecutive word sequences) for machine translation. On the other hand, BLEURT aims to achieve human-quality scoring. Higher scores indicate better translation. We use the sacreBLEU \cite{post2018call} implementation for BLEU, and BLEURT-20 checkpoints \cite{pu2021learning} for the calculation of BLEURT scores.

\subsection{Quantitative Results}
\textbf{Sign Language Recognition:} We evaluate our I3D model on the MeineDGS dataset providing the per-instance and per-class accuracy scores. We first fine-tuned I3D pretrained on Kinetics \cite{carreira2017quo} and obtained 53.24\% and 40.70\%, respectively. 
It has been shown that pretraining on larger sign recognition datasets improves sign performance \cite{albanie2020bsl, vazquez2021isolated}. Therefore, we first fine-tuned the I3D model on the large-scale BOBSL (BBC-Oxford British Sign Language) dataset \cite{albanie2021bbc}. Then we fine-tuned an I3D model pretrained on Kinetics + BOBSL for finetuning on the MeineDGS. We observe ~1.5\% performance increase with the usage of sign language data in weight initialization. Our results are provided in Table \ref{table:recognition}.

\begin{table}[!h]
\caption{Isolated sign language recognition performance on MeineDGS-V. }
 \setlength{\tabcolsep}{3.4pt} 
\label{table:recognition}
    \begin{center}
    \begin{tabular}{l|lcc}
        \hline
        Model & Pretrained on & Per-instance & Per-class \\ \hline
        I3D  & Kinetics \cite{carreira2017quo} & 53.24\% & 40.70\% \\ 
        I3D  & Kinetics \cite{carreira2017quo} + BOBSL \cite{albanie2021bbc} & \textbf{54.57}\% & \textbf{42.48}\% \\ \hline
    \end{tabular}
    \end{center}
\end{table}

\textbf{Sign Language Translation:} We evaluate our entire SLT approach, Spotter+GPT, on the MeineDGS-V test split and our DGS-20 videos. The quantitative results are provided in Table \ref{tab:meinedgs} and Table \ref{tab:demo}, respectively.

We empirically evaluate the Spotter's performance using varying probability thresholds. The threshold of 0.7 for the probability associated with each gloss prediction yielded the best results.

It is worth mentioning that DGS-20 videos result in higher performance than MeineDGS. As this dataset contains a limited and controlled vocabulary, GPT achieves significantly higher scores in both BLEU and BLEURT metrics, demonstrating its effectiveness in low-resource but constrained setups.

\textbf{Evaluation of each component.} To evaluate the performance of our components independently we conduct two different types of experiments.  First, to evaluate the performance of our spotter, we replace its results with the ground truth gloss annotations. Although the MeineDGS-V test split has 4,620 glosses, our spotter is only able to recognize 2,301 glosses. To make a fair comparison, we excluded gloss annotations that do not belong to our Spotter's vocabulary and we refer to this filtered reference set as Sub-GT. Specifically, Sub-GT is derived from the full ground truth annotations in the MeineDGS-V, but only includes glosses that exist in our predefined vocabulary. We refer to our full pipeline using this subset as Sub-GT+GPT. As expected, these scores surpass the results obtained with the spotter (BLEURT: 29.72 vs. 21.62).

Second, to evaluate the role of ChatGPT as a gloss-to-text generator, we replace it with a traditional Transformer model \cite{vaswani2017attention}, trained on gloss-sentence pairs. We use two layers with 8-heads in the transformer encoder and decoder using 512 hidden units. We use the Adam optimizer with an initial learning rate of $6 \times 10^{-4}$ with batch size 64. We reduce the learning rate by a factor of 0.7 if the BLEU-4 score does not increase for 5 epochs. On MeineDGS, the transformer outperforms GPT in terms of BLEU. However, when considering BLEURT, which reflects semantic similarity, GPT achieves better results (Table \ref{tab:meinedgs}). 

On the DGS-20 videos, we evaluate the same Transformer model that was trained on the MeineDGS dataset. ChatGPT significantly outperforms transformers in all metrics (Table \ref{tab:demo}). This performance can be attributed to these factors: the spotter achieves high accuracy due to the limited vocabulary, and ChatGPT effectively maps detected gloss sequences into spoken language sentences. These results highlight that when gloss detection is reliable, prompting an LLM like ChatGPT can produce high-quality translations without any task-specific fine-tuning.

\begin{table}[!ht]
     \caption{Performance of our approach on MeineDGS-V test set. Color-coded for easier comparison: GPT vs. Transformer \textcolor{blue}{(blue)}, Spotter vs. Sub-GT \textcolor{orange}{(orange)}.}
\centering
 \setlength{\tabcolsep}{4pt} 
\begin{tabular}{ l|ccccc}
    \hline
    \textbf{Method} &  \textbf{B-1}  & \textbf{B-2}  & \textbf{B-3}  & \textbf{B-4} & \textbf{BLEURT} \\ \hline 
     \textcolor{orange}{Spotter}+\textcolor{blue}{GPT} & 14.82 & 4.19 & 1.45 & 0.64 & \textbf{21.62}  \\ 
     Spotter+\textcolor{blue}{Transformer} &  \textbf{19.5} &  \textbf{6.13} & \textbf{2.48} & \textbf{1.08} & 19.01 \\ 
     \textcolor{orange}{Sub-GT}+GPT & 16.65 & 6.45 & 3.02 & 1.55 & 29.72 \\  
     \hline

\end{tabular}
\label{tab:meinedgs}
\end{table}

 \begin{table}[!ht]
     \caption{Performance of our approach on DGS-20. Color-coded for easier comparison: GPT vs. Transformer \textcolor{blue}{(blue)}, Spotter vs. GT \textcolor{orange}{(orange)}.}
\centering
 \setlength{\tabcolsep}{3.5pt} 
\begin{tabular}{ l|ccccc}
    \hline
    \textbf{Model} &  \textbf{B-1}  & \textbf{B-2}  & \textbf{B-3}  & \textbf{B-4} & \textbf{BLEURT} \\ \hline
    \textcolor{orange}{Spotter}+\textcolor{blue}{GPT}& \textbf{38.25} & \textbf{24.13} & \textbf{15.81} & \textbf{9.12} & \textbf{46.93}  \\ 
    Spotter+\textcolor{blue}{Transformer} &  18.65 & 4.91 & 1.79 & 0.92 &  19.85  \\ 
    \textcolor{orange}{GT}+GPT & 68.74 & 59.23	& 52.21	& 46.22 &	79.04 \\ \hline
 
\end{tabular}
\label{tab:demo}
\end{table}

\begin{table*}[!h]
 \caption{Qualitative results of the proposed method. \\
 Examples 1-3 are from MeineDGS; Examples 4-6 are from DGS-20.}
	\centering
	\begin{tabular}{p{3cm}p{12cm}c}
        \hline
        GT Glosses & NUR2B NUM-EINER1A:1d PROGRAMM1A ICH1 DDR4 ES-GIBT3 \\
        Spoken Language & Es gab nur ein Programm in der DDR. (There was only one program in the GDR.)\\   
        Spotter & NUM-EINER PROGRAMM DDR ES-GIBT \\
        Spotter+GPT& Es gibt nur ein Programm in der DDR. (There is only one program in the GDR.)  \\ \hline \hline

        GT Glosses & SEHR-VIEL2 FEIN1 ESSEN1 \\
        Spoken Language & Dort gibt es sehr gutes Essen. (There is very good food there.) \\
        Spotter & SEHR-VIEL GUT ESSEN \\
         Spotter+GPT& Es gibt sehr viel gutes Essen. (There is a lot of good food.) \\
        
        \hline \hline

        GT Glosses & LETZTE URLAUB SARDINIEN INDEX\\
        Spoken Language & Letztens war ich im Urlaub auf Sardinien. (I was recently on holiday in Sardinia.) \\
        Spotter & LETZTE MEHR  \\
         Spotter+GPT& Letzte Woche gab es mehr Regen als erwartet. (Last week there was more rain than expected.)  \\ \hline \hline

        GT Glosses & FAMILIE1 ESSEN1 ABEND2 RESTAURANT1\\
        Spoken Language & Die Familie isst abends im Restaurant. 
(The family eats in the restaurant in the evening.)  \\   
        Spotter & FAMILIE ESSEN ABEND RESTAURANT \\
        Spotter+GPT& Die Familie geht am Abend ins Restaurant, um zu essen. (The family goes to the restaurant to eat in the evening.) \\ \hline \hline

        GT Glosses & DOLMETSCHER1 SPRECHEN3 FAMILIE1 MADCHEN1 \\
        Spoken Language & Der Dolmetscher spricht mit der Familie des Mädchens. (The interpreter speaks to the girl's family.) \\   
        Spotter & DOLMETSCHER MIT MADCHEN SPRECHEN  \\
        Spotter+GPT& Der Dolmetscher spricht mit dem Mädchen. (The interpreter speaks to the girl.) \\ \hline \hline
        
        GT Glosses & DOZENT1 MORGEN1B FAHRRAD1 UNIVERSITAT1 \\
        Spoken Language & Der Dozent fährt morgens mit dem Fahrrad zur Universität. (Tomorrow the lecturer will ride his bike to the university.) \\   
        Spotter & DOZENT MORGEN FAHRRAD LUFT \\
        Spotter+GPT& Morgen fahre ich mit dem Fahrrad zur Uni, weil mein Dozent sagt, frische Luft ist wichtig. (Tomorrow I'll ride my bike to university because my lecturer says fresh air is important.)  \\ \hline \hline
        \end{tabular}
    \label{tab:examples}
	\end{table*}

\subsection{Qualitative results} 
We provide qualitative results of our approach in Table \ref{tab:examples}. When the spotter successfully detects the majority of glosses, GPT effectively generates high-quality spoken sentences from the glosses (Example 1, 2, and 4). Even in cases where the glosses are incomplete or contain additional glosses, ChatGPT still preserves semantic coherence with information gaps or the generation of new content (Examples 5, and 6). On the other hand, not surprisingly, ChatGPT’s performance heavily depends on the quality of the input glosses. When the spotter fails to detect glosses, ChatGPT generates incorrect sentences (Example 3) or ``No translation".

\section{Conclusion}
\label{sec:conclusion}
In this paper, we proposed a novel sign language translation framework that combines a sign spotter with a Large Language Model (LLM), specifically ChatGPT, to generate spoken language sentences from sign language videos. Our method does not require any end-to-end SLT model training and leverages prompt-based inference for gloss-to-text generation.

Experimental results on MeineDGS-V and a newly collected DGS-20 dataset show that Spotter+GPT produces coherent and semantically accurate sentences, especially when the spotter successfully identifies relevant glosses. This indicate that leveraging LLMs offers a promising and flexible alternative to traditional gloss-to-text pipelines.

Our system can process video inputs from both pre-recorded datasets and live capture devices such as webcams. This flexibility allows us to support potential real-time applications.

However, our system's performance is inherently constrained by the vocabulary and accuracy of the sign spotter. When the spotter fails to detect critical glosses, the translation quality drops significantly. This highlights the importance of high-quality gloss spotting. Nevertheless, our approach can be adapted to specialized sign language interpretation by fine-tuning the spotter on domain-specific gloss sets. 
A future direction may include expanding the vocabulary of the spotter to increase the range of recognized glosses.

\begin{acks}
We would like to thank Necati Cihan Camgoz for the valuable discussions and feedback. This work was supported by the SNSF project ‘SMILE II’ (CRSII5 193686), the Innosuisse IICT Flagship (PFFS-21-47), EPSRC grant APP24554 (SignGPT-EP/Z535370/1), and through funding from Google.org via the AI for Global Goals scheme. This work reflects only the author’s views and the funders are not responsible for any use that may be made of the information it contains.
\end{acks}

\bibliographystyle{ACM-Reference-Format}
\bibliography{refs}

\end{document}